%% file: iclr2021_conference.tex
\documentclass{article} 
\usepackage{iclr2021_conference,times}

\input{math_commands.tex}

\usepackage{hyperref}
\usepackage{url}
\usepackage{graphicx}
\usepackage{wrapfig}

\title{A Latent space solver for PDE generalization}

\author{Rishikesh Ranade, Chris Hill, Haiyang He, Amir Maleki \& Jay Pathak \\
Ansys Inc\\
Canonsburg, PA 15317, USA \\
\texttt{\{rishikesh.ranade, chris.hill, haiyang.he, amir.maleki \& jay.pathak\}@ansys.com} \\
}

\iclrfinalcopy 
\begin{document}

\maketitle

\begin{abstract}

In this work we propose a hybrid solver to solve partial differential equation (PDE)s in the latent space. The solver uses an iterative inferencing strategy combined with solution initialization to improve generalization of PDE solutions. The solver is tested on an engineering case and the results show that it can generalize well to several PDE conditions.

\end{abstract}
\section{Introduction}
Simulations are important in engineering applications to explore the underlying physics. But, they can be computationally expensive because they involve numerical methods to solve partial differential equations (PDEs) for various conditions associated with the PDE, such as geometry of computational domain, type of boundary conditions (BC), external source terms etc. Researchers have explored the idea of coupling machine learning with PDEs for several decades \citep{crutchfield1987equations, kevrekidis2003equation}. Recently, there is a tremendous focus on improving the predictive capability and generalizability of ML methods by infusing physics, either during training or prediction. A substantial portion of research has focused on introducing physics-based constraints in neural networks through the computation of PDE derivatives \citep{raissi2019physics, raissi2018hidden, rao2020physics, ranade2020discretizationnet, gao2020phygeonet, wu2018physics, qian2020lift, xue2020amortized}. More recently, there is a huge effort on training neural networks within frameworks of differentiable PDE solvers. These approaches use differentiable solvers to learn and control PDE solutions as well as the dynamics of the system \citep{amos2017optnet, um2020solver, de2018end, toussaint2018differentiable, wang2020differentiable, holl2020learning, portwood2019turbulence, bar2019learning, zhuang2020learned, kochkov2021machine}. 

There is a growing need for improving generalizability of ML techniques to a wider range of PDE parameters. The existing ML approaches that learn from PDE conditions, such as geometry, BCs and source terms have to handle several challenges. Firstly, the PDE conditions can be high dimensional and sparse, impacting learning and generalizability. Secondly, the inference procedure is static and there are limited opportunities to alter the trajectory of PDE solutions. Finally, the space of physical solutions is very large.

In this paper, we propose a hybrid solver to learning PDE solutions and address the outlined challenges. We use lower dimensional representations to tackle the issues of high dimensionality and sparsity of PDE conditions and solutions. Further, the inferencing methodology uses an iterative procedure to solve the PDE in a lower dimensional latent space with fixed point iterations. Latent space learning has been explored in several works, \citep{wiewel2020latent, maulik2020reduced, kim2019deep, murata2020nonlinear, fukami2020convolutional, champion2019data, fukami2020sparse, he2020unsupervised}, but our approach to solve PDEs is different and novel. The iterative procedure at inference enables us to alter the solution trajectories using existing PDE solvers. This is useful in ML based techniques to ensure solver robustness, generalizability and accuracy. The hybrid solver combines solution initialization using existing coarse grid PDE solvers with this iterative inferencing procedure and is demonstrated on a practical engineering application.

\section{Solution Methodology}
\subsection{Latent space representation}
\begin{figure}[h]
\begin{center}
    \includegraphics[scale=0.4]{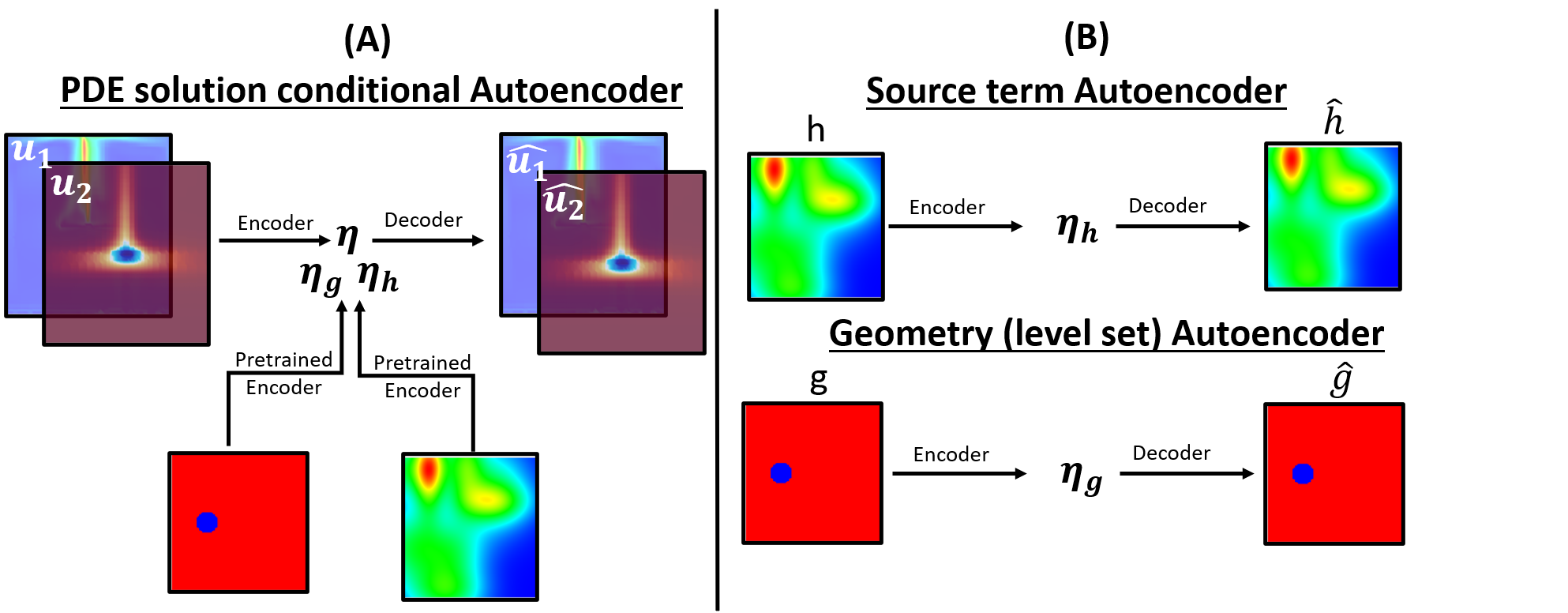}
\end{center}
\caption{Autoencoders for PDE conditions and solution}
\label{fig:1}
\end{figure}
Fig. \ref{fig:1} shows the neural network architectures used to determine the compressed latent space vectors of the various PDE conditions, such as geometry of computational domain, BCs and source term distributions, and PDE solutions ($u_1$ and $u_2$). The geometry of the computational domain is represented using a binarized level set representation \citep{osher1988fronts}. On the other hand, boundary conditions and source terms already have spatio-temporal distributions. The PDE conditions, as well as the PDE solutions are compressed into their lower dimensional latent vectors, $\eta_g, \eta_h, \eta_b, \eta$ , using CNN encoder-decoder type networks. The loss functions used to constrain the Autoencoder networks of PDE conditions are purely statistical. Conversely, the PDE solution Autoencoders are augmented by including PDE based loss constraints, similar to \citet{ranade2020discretizationnet}. The PDE constraints are computed using the discretization schemes, available in the Ansys suite of software. Additionally, since the PDE solutions are dependent on other PDE conditions, the solution Autoencoders are also conditioned upon the latent vectors of the PDE conditions. 

\subsection{Hybrid latent space solver methodology} \label{hyb_sol}
\begin{figure}[h]
\begin{center}
    \includegraphics[scale=0.4]{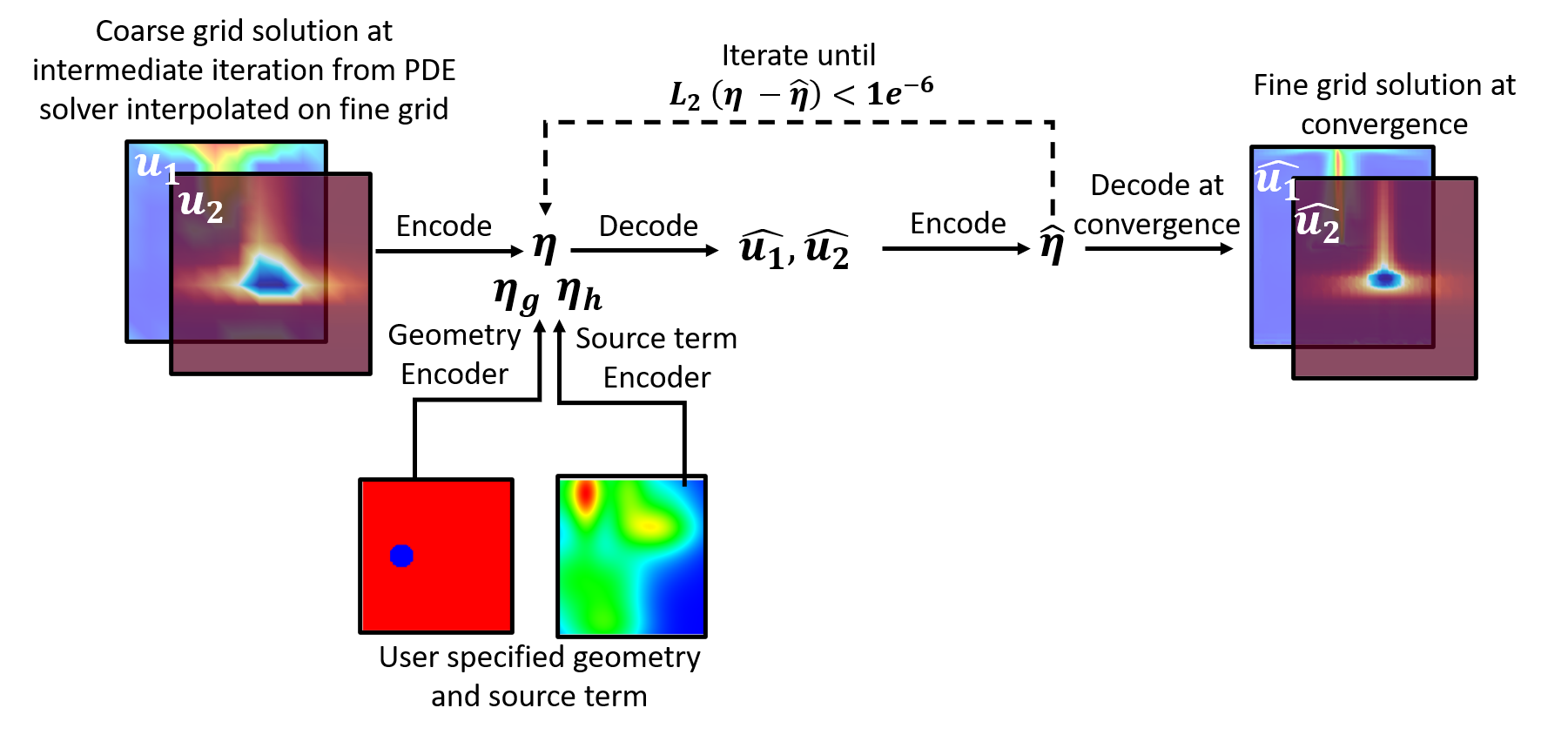}
\end{center}
\caption{Hybrid solver: Iterative inferencing strategy}
\label{fig:2}
\end{figure}
Fig. \ref{fig:2} shows the hybrid latent space solver methodology proposed in this work for using the Autoencoder networks to infer at unknown and unseen conditions. The iterative procedure based on fixed point iterations and is motivated from our previous work, \citep{ranade2020discretizationnet}. The different steps involved in the solution procedure are outlined below.

\begin{enumerate}
    \item Latent vectors, $\eta_g$, $\eta_h$ and $\eta_b$, are computed for a new computational domain geometry, boundary condition and source term distribution using the encoder networks.
    \item The initial solution of PDE is computed using an existing PDE solver on a very coarse grid. These solutions are interpolated on the fine grid and encoded to their latent vector form, $\eta$.
    \item The initial solution latent vector, $\eta$ combined with $\eta_g$, $\eta_h$ and $\eta_b$ is passed through the decoder to generate solution fields, ${u_1}, {u_2}$.
    \item Solution fields are compressed to a new solution latent vector, $\hat{\eta}$, using the the encoder.
    \item Steps $3$ and $4$ are repeated with the new solution latent vector, $\hat{\eta}$ until  $||\eta-\hat{\eta}||_2 < 1e^{-6}$. 
    \item At convergence, the PDE solutions are decoded using the most recent $\hat{\eta}$.
\end{enumerate}

The hybrid latent space solver described above has two main implications. Firstly, the iterative procedure used for inferencing allows initialization of solutions using existing PDE solvers and moreover provides an opportunity to intervene the solution process and alter the solution trajectory using PDE solvers. Secondly, the solution procedure is conditioned by richer, lower dimensional representations of PDE conditions in the form of latent vectors, thus enhancing the generalizability.

\section{Results and discussions}
\begin{wrapfigure}{r}{0.47\textwidth}
    \centering
    \vspace{-1em}
    \includegraphics[width=6cm]{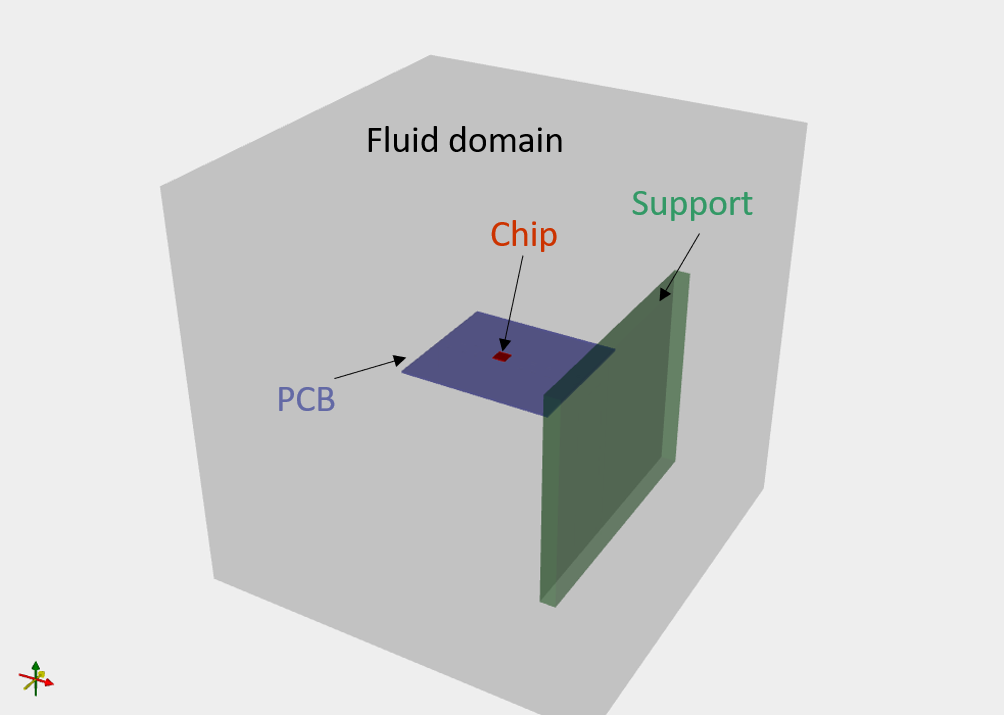}
    \vspace{-1em}
    \caption{Natural convection of electronic chip}
    \vspace{-1em}
    \label{fig:3}
\end{wrapfigure}
The hybrid solver is demonstrated for a 3-D, steady-state electronic cooling case with natural convection. There are $5$ solution variables, $3$ components of velocity, pressure and temperature. The geometry of the computational domain may be observed in Fig. \ref{fig:3}. The domain consists of a chip-mold assembly held by a PCB and the entire geometry is placed inside a fluid domain. The chip is subjected to heat sources with random spatial distributions due to uncertainty in electrical heating. Fig. \ref{fig:4} in Appendix \ref{App} shows an example of the $8$ different distributions of heat sources. From a physics standpoint, natural convection results in a two-way coupling between temperature and velocity. The heat source specified results in a temperature increase, which generates fluid velocity because of buoyancy effects and in turn cools the chip. At steady state there is sufficient velocity generation to cancel out the heat generation. The main challenges in this case are to capture the complicated physics resulting from the two-way coupling and to generalize to different spatial distributions of heat sources.

\subsection{Data generation and training}
The Autoencoder for heat source distribution is trained with random spatial distributions generated using a Gaussian mixture model, where the number of Gaussians are varied from $1$ to $20$ for randomized mean and variance. The Autoencoder is set up to achieve a compression ratio of around $12$. The PDE solution Autoencoder is trained with $200$ solution generated using Ansys Fluent fluid flow simulation software on a computational mesh with $128^3$ elements for random heat sources. A compression ratio of $64$ is achieved. The normalized reconstruction mean squared error for unseen test samples is to the order of $1e^{-6}$ for all the solution variables as well as the heat source.

\subsection{Comparisons with Ansys Fluent}

The hybrid solver is compared with Ansys Fluent solution for two heat source distributions, which are generated randomly and thus, never seen by the networks. The initial solution is computed at a mesh resolution of $16$ elements in each spatial direction in Ansys Fluent for 200 steady-state iterations.

\subsubsection{Computation time}
Ansys Fluent takes an average of $200$ minutes on a single CPU to obtain a converged solution on $128^3$ mesh. On the other hand, the hybrid solver converges in $50$ seconds on average, including the time it takes to generate the coarse grid solution. Thus, the hybrid solver results in a 200x speedup over Ansys Fluent in generating solutions on fine girds. The averages are calculated over runs on 100 cases with different heat source distributions.

\subsubsection{Contour \& Line plots}
Fig. \ref{fig:5} and Fig. \ref{fig:6} in Appendix \ref{App} compare the solution contours on the X-Y plane for the two test cases. The hybrid solution agrees well with respect to the Ansys Fluent solutions with small mean squared errors. Although not shown here, the performance is similar for a variety of random distributions of heat source, including the examples shown in Figure \ref{fig:4} in the Appendix \ref{App}. The line plots in Fig. \ref{fig:7} and Fig. \ref{fig:8} in Appendix \ref{App} are plotted along the Y direction through chip center. It can be observed that the overall trends as well as the peak quantities agree well for both velocity and temperature in both test cases. The contour and line plots show that the hybrid latent space solver captures the two-way coupled physics accurately and more importantly, the methodology generalizes well to a variety of heat source distributions.

\begin{figure}[h]
\begin{center}
    \includegraphics[scale=0.25]{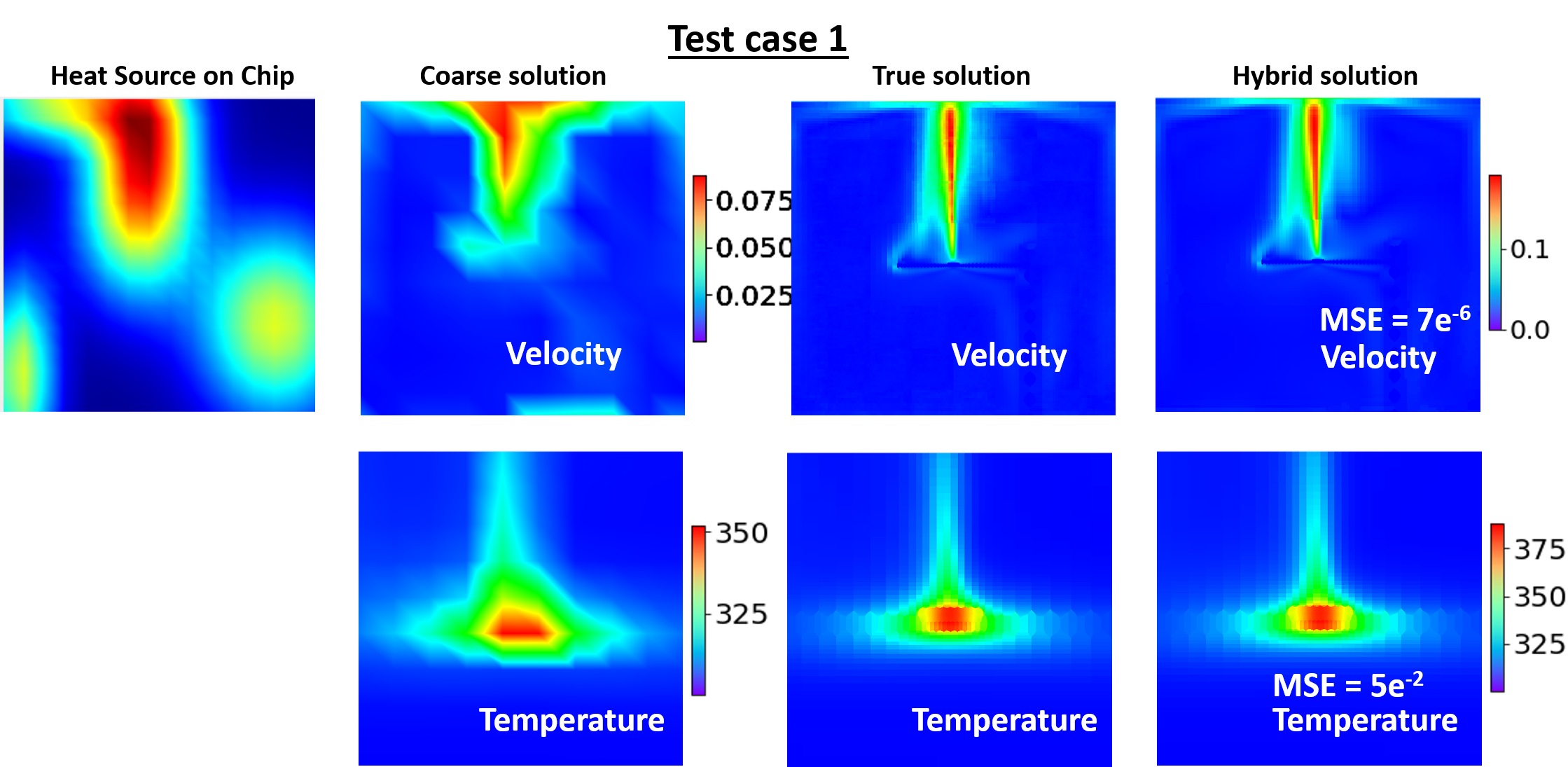}
\end{center}
\caption{Comparisons of velocity and temperature between hybrid solver solution and Ansys Fluent solution for test case 1. The contours are plotted in X-Y plane perpendicular to the chip. The velocity contour represents the entire domain (X$\in$(-0.1525m to 0.1525m), Y$\in$(-0.1525m, 0.1525m)), while the temperature contour zooms on the chip (X$\in$(-0.03m, 0.03m), Y$\in$(-0.03m, 0.03m).}
\label{fig:5}
\end{figure}

\begin{figure}[h]
\begin{center}
    \includegraphics[scale=0.3]{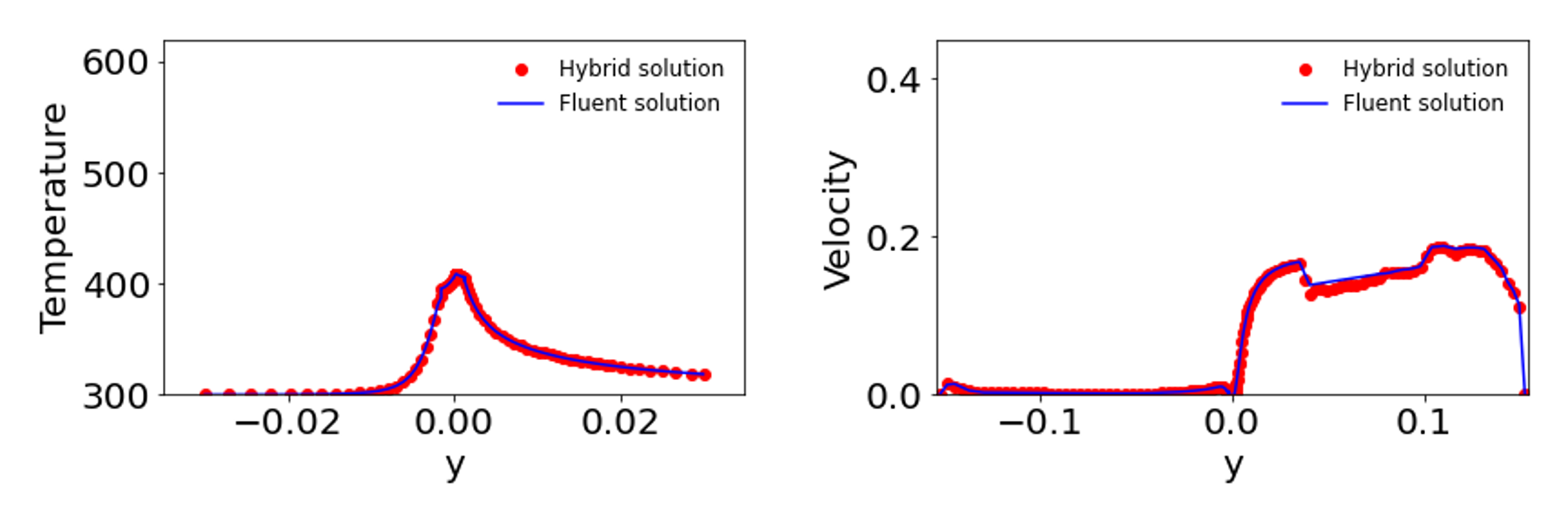}
\end{center}
\caption{Comparisons of velocity and temperature between hybrid solver solution and Ansys Fluent solution for test case 1. The line plots are plotted along Y direction at (X, Z)$\in$(0.0m, 0.0m). The velocity line plot represents the entire domain (Y$\in$(-0.1525m, 0.1525m)), while the temperature line plot zooms on the chip (Y$\in$(-0.03m, 0.03m).}
\label{fig:7}
\end{figure}

\section{Conclusion and Future work}

In this work we have proposed a hybrid solver that combines latent space learning with solution initialization to improve generalization of PDE solutions. The results shows that the methodology is computationally fast, accurate and generalizable. In its current form, the proposed methodology has several limitations. Most notably, the approach is restricted to structured domains with fixed resolutions with relatively simpler computational domain geometries. In future, we would like to extend this to solve on unstructured meshes. Furthermore, the latent space iterative strategy will be integrated with an actual PDE solver to alter solution trajectories and improve convergence.   

\bibliography{iclr2021_conference}
\bibliographystyle{iclr2021_conference}

\newpage
\appendix
\section{Appendix} \label{App}

\begin{figure}[h]
\begin{center}
    \includegraphics[scale=0.25]{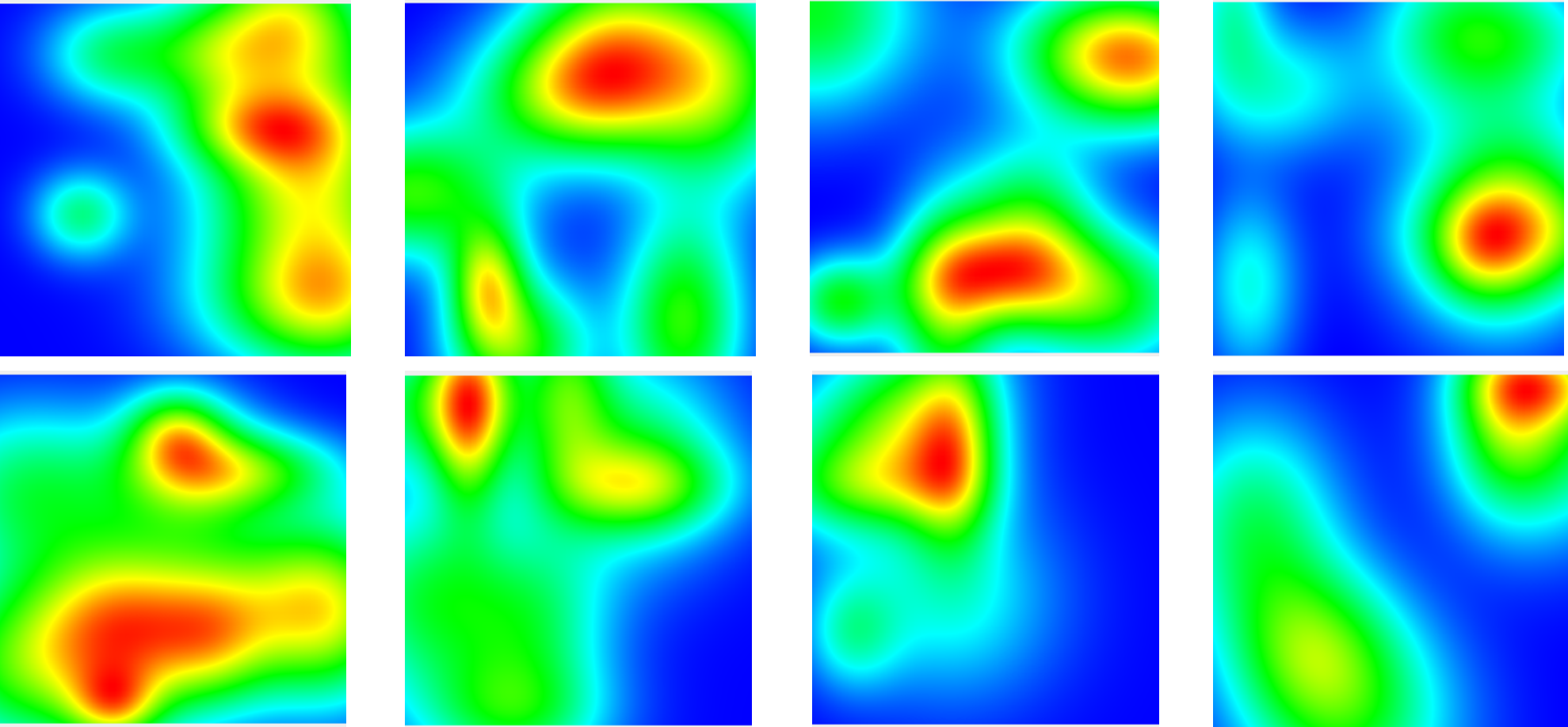}
\end{center}
\caption{Example of randomness in spatial distribution of heat source terms}
\label{fig:4}
\end{figure}

\begin{figure}[h]
\begin{center}
    \includegraphics[scale=0.25]{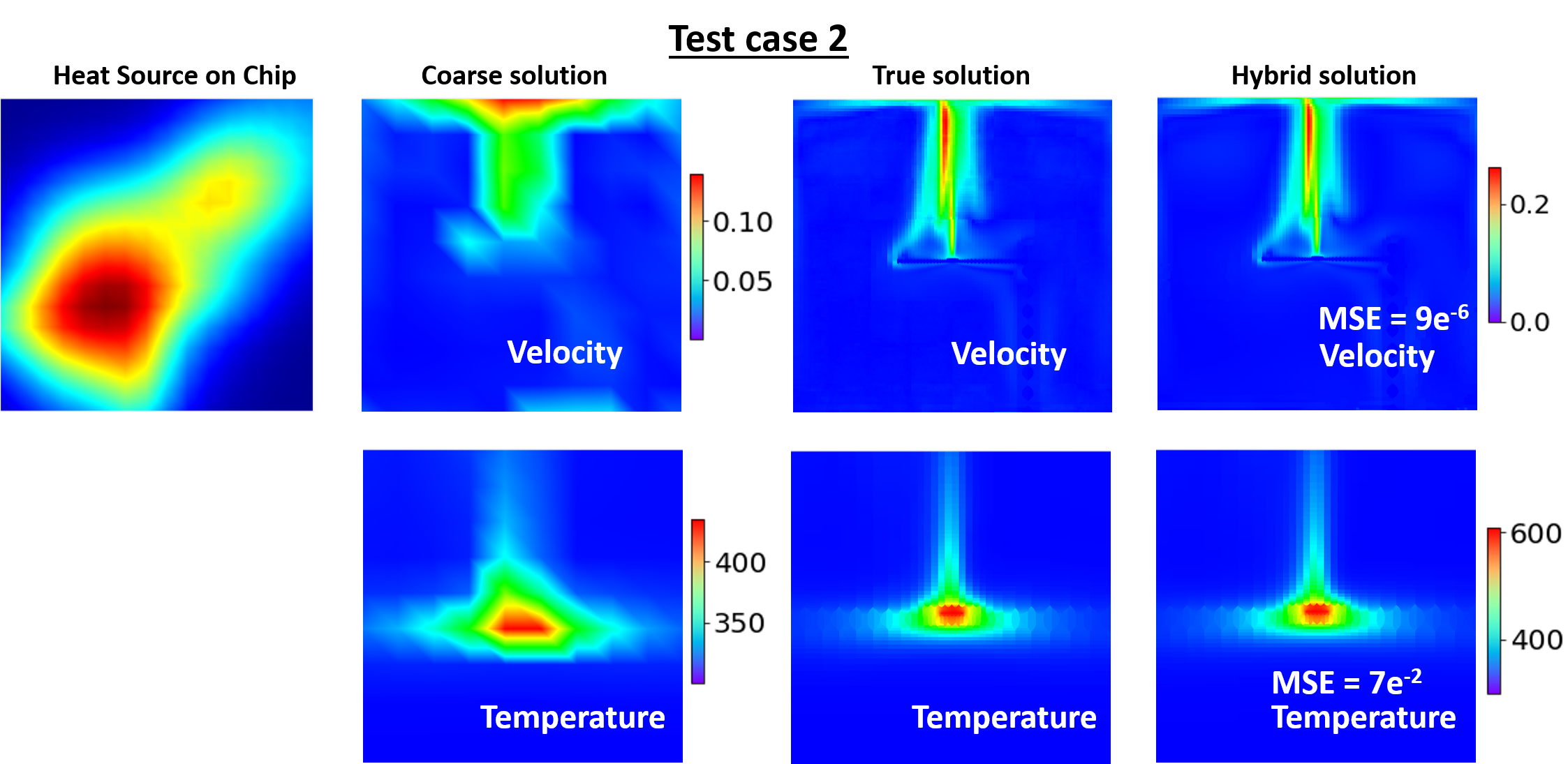}
\end{center}
\caption{Comparisons of velocity and temperature between hybrid solver solution and Ansys Fluent solution for test case 2. The contours are plotted in X-Y plane perpendicular to the chip. The velocity contour represents the entire domain (X=(-0.1525m to 0.1525m), Y$\in$(-0.1525m, 0.1525m)), while the temperature contour zooms on the chip (X$\in$(-0.03m, 0.03m), Y=(-0.03m, 0.03m).}
\label{fig:6}
\end{figure}

\begin{figure}[h]
\begin{center}
    \includegraphics[scale=0.3]{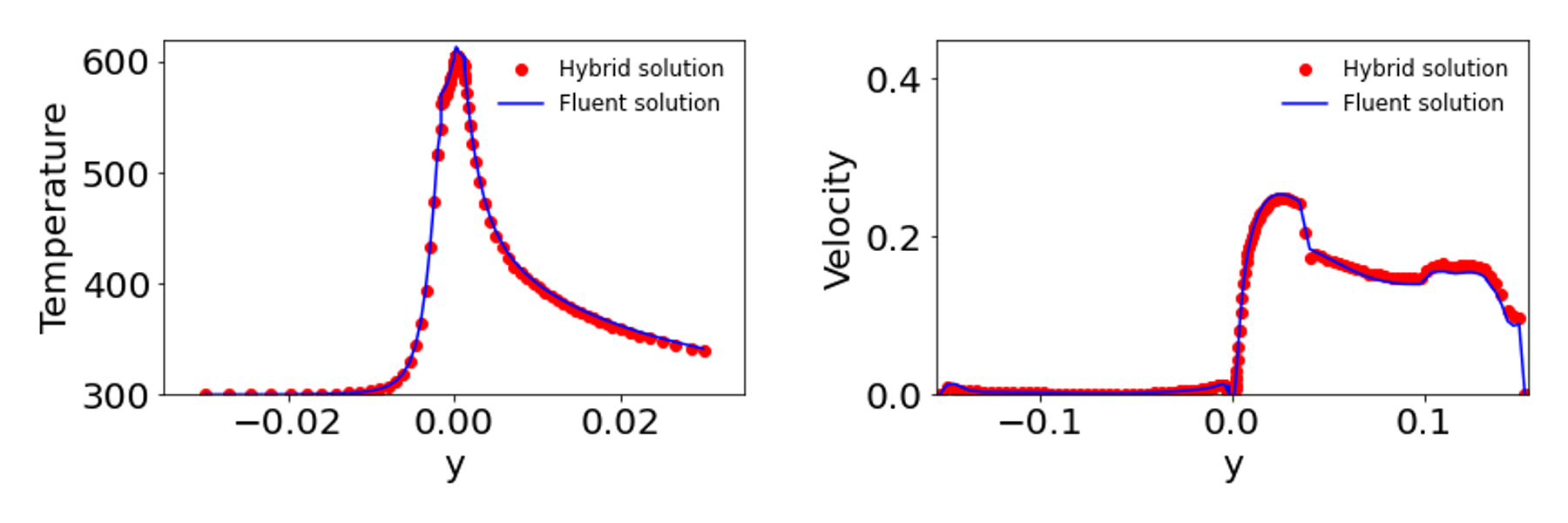}
\end{center}
\caption{Comparisons of velocity and temperature between hybrid solver solution and Ansys Fluent solution for test case 2. The line plots are plotted along Y direction at (X, Z)$\in$0.0m, 0.0m). The velocity line plot represents the entire domain (Y$\in$(-0.1525m, 0.1525m)), while the temperature line plot zooms on the chip (Y=(-0.03m, 0.03m).}
\label{fig:8}
\end{figure}

\end{document}

%% file: math_commands.tex
\usepackage{amsmath,amsfonts,bm}

\def\eqref#1{equation~\ref{#1}}

\def\1{\bm{1}}

\DeclareMathAlphabet{\mathsfit}{\encodingdefault}{\sfdefault}{m}{sl}
\SetMathAlphabet{\mathsfit}{bold}{\encodingdefault}{\sfdefault}{bx}{n}













%% file: iclr2021_conference.bbl
\begin{thebibliography}{29}
\providecommand{\natexlab}[1]{#1}
\providecommand{\url}[1]{\texttt{#1}}
\expandafter\ifx\csname urlstyle\endcsname\relax
  \providecommand{\doi}[1]{doi: #1}\else
  \providecommand{\doi}{doi: \begingroup \urlstyle{rm}\Url}\fi

\bibitem[Amos \& Kolter(2017)Amos and Kolter]{amos2017optnet}
Brandon Amos and J~Zico Kolter.
\newblock Optnet: Differentiable optimization as a layer in neural networks.
\newblock In \emph{International Conference on Machine Learning}, pp.\
  136--145. PMLR, 2017.

\bibitem[Bar-Sinai et~al.(2019)Bar-Sinai, Hoyer, Hickey, and
  Brenner]{bar2019learning}
Yohai Bar-Sinai, Stephan Hoyer, Jason Hickey, and Michael~P Brenner.
\newblock Learning data-driven discretizations for partial differential
  equations.
\newblock \emph{Proceedings of the National Academy of Sciences}, 116\penalty0
  (31):\penalty0 15344--15349, 2019.

\bibitem[Champion et~al.(2019)Champion, Lusch, Kutz, and
  Brunton]{champion2019data}
Kathleen Champion, Bethany Lusch, J~Nathan Kutz, and Steven~L Brunton.
\newblock Data-driven discovery of coordinates and governing equations.
\newblock \emph{Proceedings of the National Academy of Sciences}, 116\penalty0
  (45):\penalty0 22445--22451, 2019.

\bibitem[Crutchfield \& McNamara(1987)Crutchfield and
  McNamara]{crutchfield1987equations}
James~P Crutchfield and BS~McNamara.
\newblock Equations of motion from a data series.
\newblock \emph{Complex systems}, 1\penalty0 (417-452):\penalty0 121, 1987.

\bibitem[de~Avila Belbute-Peres et~al.(2018)de~Avila Belbute-Peres, Smith,
  Allen, Tenenbaum, and Kolter]{de2018end}
Filipe de~Avila Belbute-Peres, Kevin Smith, Kelsey Allen, Josh Tenenbaum, and
  J~Zico Kolter.
\newblock End-to-end differentiable physics for learning and control.
\newblock \emph{Advances in neural information processing systems},
  31:\penalty0 7178--7189, 2018.

\bibitem[Fukami et~al.(2020{\natexlab{a}})Fukami, Murata, and
  Fukagata]{fukami2020sparse}
Kai Fukami, Takaaki Murata, and Koji Fukagata.
\newblock Sparse identification of nonlinear dynamics with low-dimensionalized
  flow representations.
\newblock \emph{arXiv preprint arXiv:2010.12177}, 2020{\natexlab{a}}.

\bibitem[Fukami et~al.(2020{\natexlab{b}})Fukami, Nakamura, and
  Fukagata]{fukami2020convolutional}
Kai Fukami, Taichi Nakamura, and Koji Fukagata.
\newblock Convolutional neural network based hierarchical autoencoder for
  nonlinear mode decomposition of fluid field data.
\newblock \emph{Physics of Fluids}, 32\penalty0 (9):\penalty0 095110,
  2020{\natexlab{b}}.

\bibitem[Gao et~al.(2020)Gao, Sun, and Wang]{gao2020phygeonet}
Han Gao, Luning Sun, and Jian-Xun Wang.
\newblock Phygeonet: Physics-informed geometry-adaptive convolutional neural
  networks for solving parametric pdes on irregular domain.
\newblock \emph{arXiv preprint arXiv:2004.13145}, 2020.

\bibitem[He \& Pathak(2020)He and Pathak]{he2020unsupervised}
Haiyang He and Jay Pathak.
\newblock An unsupervised learning approach to solving heat equations on chip
  based on auto encoder and image gradient.
\newblock \emph{arXiv preprint arXiv:2007.09684}, 2020.

\bibitem[Holl et~al.(2020)Holl, Koltun, and Thuerey]{holl2020learning}
Philipp Holl, Vladlen Koltun, and Nils Thuerey.
\newblock Learning to control pdes with differentiable physics.
\newblock \emph{arXiv preprint arXiv:2001.07457}, 2020.

\bibitem[Kevrekidis et~al.(2003)Kevrekidis, Gear, Hyman, Kevrekidid, Runborg,
  Theodoropoulos, et~al.]{kevrekidis2003equation}
Ioannis~G Kevrekidis, C~William Gear, James~M Hyman, Panagiotis~G Kevrekidid,
  Olof Runborg, Constantinos Theodoropoulos, et~al.
\newblock Equation-free, coarse-grained multiscale computation: Enabling
  mocroscopic simulators to perform system-level analysis.
\newblock \emph{Communications in Mathematical Sciences}, 1\penalty0
  (4):\penalty0 715--762, 2003.

\bibitem[Kim et~al.(2019)Kim, Azevedo, Thuerey, Kim, Gross, and
  Solenthaler]{kim2019deep}
Byungsoo Kim, Vinicius~C Azevedo, Nils Thuerey, Theodore Kim, Markus Gross, and
  Barbara Solenthaler.
\newblock Deep fluids: A generative network for parameterized fluid
  simulations.
\newblock In \emph{Computer Graphics Forum}, volume~38, pp.\  59--70. Wiley
  Online Library, 2019.

\bibitem[Kochkov et~al.(2021)Kochkov, Smith, Alieva, Wang, Brenner, and
  Hoyer]{kochkov2021machine}
Dmitrii Kochkov, Jamie~A Smith, Ayya Alieva, Qing Wang, Michael~P Brenner, and
  Stephan Hoyer.
\newblock Machine learning accelerated computational fluid dynamics.
\newblock \emph{arXiv preprint arXiv:2102.01010}, 2021.

\bibitem[Maulik et~al.(2020)Maulik, Lusch, and Balaprakash]{maulik2020reduced}
Romit Maulik, Bethany Lusch, and Prasanna Balaprakash.
\newblock Reduced-order modeling of advection-dominated systems with recurrent
  neural networks and convolutional autoencoders.
\newblock \emph{arXiv preprint arXiv:2002.00470}, 2020.

\bibitem[Murata et~al.(2020)Murata, Fukami, and Fukagata]{murata2020nonlinear}
Takaaki Murata, Kai Fukami, and Koji Fukagata.
\newblock Nonlinear mode decomposition with convolutional neural networks for
  fluid dynamics.
\newblock \emph{Journal of Fluid Mechanics}, 882, 2020.

\bibitem[Osher \& Sethian(1988)Osher and Sethian]{osher1988fronts}
Stanley Osher and James~A Sethian.
\newblock Fronts propagating with curvature-dependent speed: algorithms based
  on hamilton-jacobi formulations.
\newblock \emph{Journal of computational physics}, 79\penalty0 (1):\penalty0
  12--49, 1988.

\bibitem[Portwood et~al.(2019)Portwood, Mitra, Ribeiro, Nguyen, Nadiga, Saenz,
  Chertkov, Garg, Anandkumar, Dengel, et~al.]{portwood2019turbulence}
Gavin~D Portwood, Peetak~P Mitra, Mateus~Dias Ribeiro, Tan~Minh Nguyen,
  Balasubramanya~T Nadiga, Juan~A Saenz, Michael Chertkov, Animesh Garg, Anima
  Anandkumar, Andreas Dengel, et~al.
\newblock Turbulence forecasting via neural ode.
\newblock \emph{arXiv preprint arXiv:1911.05180}, 2019.

\bibitem[Qian et~al.(2020)Qian, Kramer, Peherstorfer, and
  Willcox]{qian2020lift}
Elizabeth Qian, Boris Kramer, Benjamin Peherstorfer, and Karen Willcox.
\newblock Lift \& learn: Physics-informed machine learning for large-scale
  nonlinear dynamical systems.
\newblock \emph{Physica D: Nonlinear Phenomena}, 406:\penalty0 132401, 2020.

\bibitem[Raissi \& Karniadakis(2018)Raissi and Karniadakis]{raissi2018hidden}
Maziar Raissi and George~Em Karniadakis.
\newblock Hidden physics models: Machine learning of nonlinear partial
  differential equations.
\newblock \emph{Journal of Computational Physics}, 357:\penalty0 125--141,
  2018.

\bibitem[Raissi et~al.(2019)Raissi, Perdikaris, and
  Karniadakis]{raissi2019physics}
Maziar Raissi, Paris Perdikaris, and George~E Karniadakis.
\newblock Physics-informed neural networks: A deep learning framework for
  solving forward and inverse problems involving nonlinear partial differential
  equations.
\newblock \emph{Journal of Computational Physics}, 378:\penalty0 686--707,
  2019.

\bibitem[Ranade et~al.(2020)Ranade, Hill, and
  Pathak]{ranade2020discretizationnet}
Rishikesh Ranade, Chris Hill, and Jay Pathak.
\newblock Discretizationnet: A machine-learning based solver for navier-stokes
  equations using finite volume discretization.
\newblock \emph{arXiv preprint arXiv:2005.08357}, 2020.

\bibitem[Rao et~al.(2020)Rao, Sun, and Liu]{rao2020physics}
Chengping Rao, Hao Sun, and Yang Liu.
\newblock Physics-informed deep learning for incompressible laminar flows.
\newblock \emph{Theoretical and Applied Mechanics Letters}, 10\penalty0
  (3):\penalty0 207--212, 2020.

\bibitem[Toussaint et~al.(2018)Toussaint, Allen, Smith, and
  Tenenbaum]{toussaint2018differentiable}
Marc~A Toussaint, Kelsey~Rebecca Allen, Kevin~A Smith, and Joshua~B Tenenbaum.
\newblock Differentiable physics and stable modes for tool-use and manipulation
  planning.
\newblock 2018.

\bibitem[Um et~al.(2020)Um, Holl, Brand, Thuerey, et~al.]{um2020solver}
Kiwon Um, Philipp Holl, Robert Brand, Nils Thuerey, et~al.
\newblock Solver-in-the-loop: Learning from differentiable physics to interact
  with iterative pde-solvers.
\newblock \emph{arXiv preprint arXiv:2007.00016}, 2020.

\bibitem[Wang et~al.(2020)Wang, Axelrod, and
  G{\'o}mez-Bombarelli]{wang2020differentiable}
Wujie Wang, Simon Axelrod, and Rafael G{\'o}mez-Bombarelli.
\newblock Differentiable molecular simulations for control and learning.
\newblock \emph{arXiv preprint arXiv:2003.00868}, 2020.

\bibitem[Wiewel et~al.(2020)Wiewel, Kim, Azevedo, Solenthaler, and
  Thuerey]{wiewel2020latent}
Steffen Wiewel, Byungsoo Kim, Vinicius~C Azevedo, Barbara Solenthaler, and Nils
  Thuerey.
\newblock Latent space subdivision: stable and controllable time predictions
  for fluid flow.
\newblock In \emph{Computer Graphics Forum}, volume~39, pp.\  15--25. Wiley
  Online Library, 2020.

\bibitem[Wu et~al.(2018)Wu, Xiao, and Paterson]{wu2018physics}
Jin-Long Wu, Heng Xiao, and Eric Paterson.
\newblock Physics-informed machine learning approach for augmenting turbulence
  models: A comprehensive framework.
\newblock \emph{Physical Review Fluids}, 3\penalty0 (7):\penalty0 074602, 2018.

\bibitem[Xue et~al.(2020)Xue, Beatson, Adriaenssens, and
  Adams]{xue2020amortized}
Tianju Xue, Alex Beatson, Sigrid Adriaenssens, and Ryan Adams.
\newblock Amortized finite element analysis for fast pde-constrained
  optimization.
\newblock In \emph{International Conference on Machine Learning}, pp.\
  10638--10647. PMLR, 2020.

\bibitem[Zhuang et~al.(2020)Zhuang, Kochkov, Bar-Sinai, Brenner, and
  Hoyer]{zhuang2020learned}
Jiawei Zhuang, Dmitrii Kochkov, Yohai Bar-Sinai, Michael~P Brenner, and Stephan
  Hoyer.
\newblock Learned discretizations for passive scalar advection in a 2-d
  turbulent flow.
\newblock \emph{arXiv preprint arXiv:2004.05477}, 2020.

\end{thebibliography}
